\documentclass[letterpaper, 10 pt, conference]{ieeeconf}  

\usepackage{booktabs}
\usepackage{graphicx}
\usepackage{amsmath}
\usepackage{amssymb}
\usepackage{hyperref}
\usepackage{cleveref}
\usepackage{float}
\usepackage{placeins}
\usepackage{listings}
\usepackage{pbalance}
\usepackage[dvipsnames]{xcolor}
\usepackage{cuted}
\usepackage{subcaption}
\usepackage{stfloats}
\usepackage{multirow}
\usepackage[absolute,overlay]{textpos}
\usepackage[style=ieee,
            doi=false,
            url=false,
            mincitenames=1,
            maxcitenames=1,
            minbibnames=6,
            maxbibnames=6,
            backend=biber]{biblatex}  
\IEEEoverridecommandlockouts 
\title{\LARGE \bf
SceneMotion: From Agent-Centric Embeddings to Scene-Wide Forecasts
}

\author{Royden Wagner$^{1}$, Ömer \c{S}ahin Ta\c{s}$^{2}$, Marlon Steiner$^{1}$, Fabian Konstantinidis$^{3}$, Hendrik Königshof$^{2}$, \\ Marvin Klemp$^{1}$, Carlos Fernandez$^{1}$, and Christoph Stiller$^{1}$
\thanks{$^{1}$Karlsruhe Institute of Technology, Engler-Bunte-Ring 21, 76131 Karlsruhe, Germany {\tt \{firstname.lastname\}@kit.edu}}%
\thanks{$^{2}$FZI Research Center for Information Technology, Haid-und-Neu-Str. 10-14, 76131 Karlsruhe, Germany {\tt \{tas, koenigshof\}@fzi.de}}%
\thanks{$^{3}$\mbox{Pre-Development of Automated Driving, CARIAD SE}, 38440 Wolfsburg, Germany {\tt fabian.konstantinidis@cariad.technology}}%
}

\begin{document}

\maketitle

\begin{strip}
\begin{minipage}{\textwidth}\centering
\vspace{-20mm}
\begin{center}
    \centering
    \label{fig:1}
    \captionsetup{type=figure}
    \includegraphics[width=\linewidth]{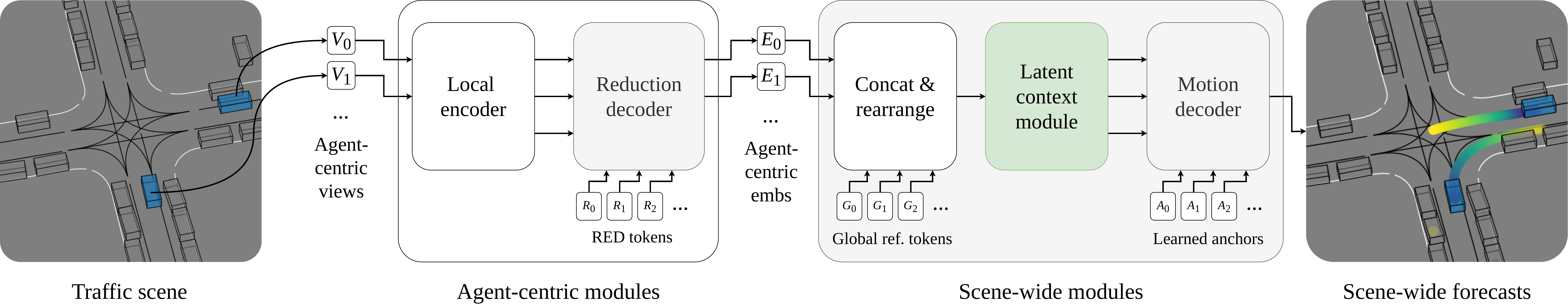}
    \captionof{figure}{\textbf{SceneMotion.} Our attention-based motion forecasting model is composed of stacked encoder and decoder modules. Variable-sized agent-centric views $V_i$ are reduced to fixed-sized agent-centric embeddings $E_i$ via cross-attention with road environment descriptor (RED) tokens $R_j$. Afterwards, we concatenate the agent-centric embeddings with global reference tokens $G_i$ and rearrange them to form a scene-wide embedding. Our latent context module then learns global context and our motion decoder transforms learned anchors $A_k$ into scene-wide forecasts. We show a simplified example with only two focal agents. By default, our model forecasts motion for 8 focal agents each with 48 context agents, enabling interaction modeling in complex scenarios.}
\end{center}%
\end{minipage}
\end{strip}

\thispagestyle{empty}
\pagestyle{empty}

\begin{abstract}
Self-driving vehicles rely on multimodal motion forecasts to effectively interact with their environment and plan safe maneuvers.
We introduce SceneMotion, an attention-based model for forecasting scene-wide motion modes of multiple traffic agents.
Our model transforms local agent-centric embeddings into scene-wide forecasts using a novel latent context module.
This module learns a scene-wide latent space from multiple agent-centric embeddings, enabling joint forecasting and interaction modeling.
The competitive performance in the Waymo Open Interaction Prediction Challenge demonstrates the effectiveness of our approach.
Moreover, we cluster future waypoints in time and space to quantify the interaction between agents.
We merge all modes and analyze each mode independently to determine which clusters are resolved through interaction or result in conflict.
Our implementation is available at: \href{https://github.com/kit-mrt/future-motion}{https://github.com/kit-mrt/future-motion} 

\end{abstract}

\section{Introduction}
\label{sec:introduction}
Future motion in traffic scenarios is inherently multimodal.
Multiple types of agents such as vehicles, pedestrians, or cyclists can move and interact in multiple ways.
Therefore, self-driving vehicles rely on multimodal motion forecasts to successfully interact with their environment and plan safe maneuvers \cite{tas2023decisiontheoretic, nishimura2023rap}.
Motion forecasting methods predict future motion based on past observations and scene context (e.g., lane data and traffic light states).
However, the majority of recent methods (e.g., \cite{nayakanti2023wayformer, wang2023prophnet}) forecast motion for each agent individually (i.e., marginal prediction), which omits modeling of future interaction between agents.

In this work, we present SceneMotion, an attention-based model for forecasting scene-wide motion modes of multiple traffic agents.
Our model transforms local agent-centric embeddings into scene-wide forecasts using a novel latent context module.
We use agent-centric representations since they allow for more data-efficient learning than scene-centric representations (as in e.g., \cite{ngiam2022scene}).
In particular, they generate more training samples (potentially for each agent in a traffic scene) and share the local reference frame with motion forecasts.
In contrast to fully agent-centric models (e.g., \cite{nayakanti2023wayformer, shi2022motion}), we use local agent-centric views that capture information from a smaller area of the map. 
Afterwards, our latent context module learns a scene-wide latent space from multiple agent-centric embeddings, enabling joint forecasting and interaction modeling.

Recent methods address interpretability in motion forecasting by engineering learned representations \cite{tas2024words, itkina2023interpretable} or learning intention goals \cite{mangalam2021goals, gilles2022thomas}.
We interpret scene-wide motion forecasts in terms of clusters of waypoints to quantify the interaction between agents. 
Waypoint clusters are related to implicit occupancy forecasts \cite{agro2023implicit, mahjourian2022occupancy}, but explicitly represent which agents are likely to interact. 
This expands the contextual information available to motion planners.

Overall, our main contributions are twofold:
\begin{enumerate}
    \item We present SceneMotion, an attention-based and data-efficient model for forecasting scene-wide motion modes.
    \item We propose to analyze scene-wide motion forecasts in terms of clusters of waypoints to quantify future interaction.
\end{enumerate}

\section{Related work}
\textbf{Joint motion forecasting} methods model the joint distribution of future motions of multiple agents.
Related methods use attention mechanisms between latent representations to model interactions and decode joint futures \cite{mercat2020multi, ngiam2022scene, girgis2022latent, steiner2024mapformer}.
Seff et al. \cite{seff2023motionlm} cast motion forecasting as language modeling and autoregressively decode joint sets of discrete motion vectors.
Their autoregressive decoding allows for temporal causal conditioning, but leads to high inference latency.
Jiang et al. \cite{jiang2023motiondiffuser} reformulate joint motion forecasting as conditional denoising diffusion process.
They condition a learned denoiser on environment context to transform a set of noisy trajectories (i.e., positions disturbed by Gaussian noise) into a joint set of future trajectories.
At inference, trajectories are sampled from pure noise, causing high inference latency.

\textbf{Conditional motion forecasting} methods forecast future motion for one agent conditioned on the forecast for one \cite{salzmann2020trajectron++, tolstaya2021identifying} or multiple other agents \cite{tang2019multiple, casas2020implicit,wirth2023conditional}.
For scene-wide forecasts, iterative rollouts are used to successively condition forecasts on each other, resulting in high inference latency for complex traffic scenarios.
However, goal conditioned motion forecasting \cite{ngiam2022scene, deo2020trajectory} bridges the gap to planning and relaxes the conditioning from a full future trajectory to a goal point (i.e., destination).

\textbf{Marginal motion forecasting} methods model future motion for each agent individually. 
Related methods extend marginal forecasting models with auxiliary joint \cite{luo2023jfp, sun2022m2i} or dense prediction \cite{gilles2022thomas} objectives to model future interactions in a second step.
However, this limits interaction modeling to recombining otherwise independent forecasts.

\textbf{Data-efficient learning} is about learning good representations from limited data.
For motion forecasting, a recent approach to data efficient learning is self-supervised learning \cite{xu2022pretram, chen2023traj, wagner2024jointmotion}.
These methods perform self-supervised pre-training followed by supervised fine-tuning using less annotated or the same data to improve downstream performance (i.e., motion forecasting accuracy).

\section{Method}
\label{sec:method}
Our proposed model, SceneMotion, is attention-based and forecasts scene-wide motion modes in self-driving applications (see Fig.\,1).
We define scene-wide motion forecasting as modeling joint distributions of future trajectories for multiple agents from past trajectories and scene context (e.g., lanes and traffic light states).
Hence, a scene-wide motion mode describes a joint set of future trajectories with one trajectory for each modeled agent.
A trajectory represents a sequence of waypoints and a waypoint is a 2D position with $x$ and $y$ coordinates.
Therefore, scene-wide motion forecasting requires scene-wide representations to model future motion for multiple agents, including interaction.

Related work focuses on improving the computational efficiency by using scene-centric \cite{ngiam2022scene, girgis2022latent} or pairwise relative representations \cite{cui2023gorela, zhang2023real}.  
Such representations reduce the amount of repeated computations compared to agent-centric input representations. 
We argue that both approaches are less data-efficient than agent-centric input representations. 
Scene-centric methods encode all agent and map data in a common reference frame, generating only one sample per scene versus one sample for each agent.
Furthermore, motion forecasts are typically in an agent-centric reference frame starting at $(x,y) = (0,0)$, which requires scene-centric models to learn the transformation from a scene-centric to an agent-centric reference frame.
Pairwise relative methods encode map data once per scene and agent history for each agent. 
As a result, agent representations are learned in a data-efficient manner, while map representations are not.

While agent-centric representations offer improved data efficiency, they can lead to redundant computations due to overlapping data.
Our model addresses this using local input views, which cover a smaller map area than in fully agent-centric models (e.g., \cite{nayakanti2023wayformer, shi2022motion, wang2023prophnet}).
Afterwards, our latent context module learns a scene-wide latent space from multiple agent-centric embeddings, enabling joint forecasting and interaction modeling.
In the following, we discuss specific representations and modules in our model in more detail.

\textbf{Input representation:} We follow \cite{gao2020vectornet, zhang2023real} and represent all processed modalities (i.e., past motion, map data, and traffic light states) as polylines.
For past motion, these polylines contain spatiotemporal data: past positions, past orientations, past velocities, dimensions, and temporal order as one-hot encodings.
We sample temporal features with a frequency of 10\,Hz.
For lane data and traffic lights, the polyline representations include node positions and classes (e.g., lane types and current traffic light states).
We generate local agent-centric input views ($V_i$ in Fig.\,1, where $i$ is the agent index) including the 128 nearest map polylines and 48 other context agents. 
For reference, the fully agent-centric Wayformer model \cite{nayakanti2023wayformer} and the pairwise relative HPTR model \cite{zhang2023real} process the nearest 512 map polylines.

\textbf{Agent-centric modules:} Our local encoder is a multilayer perceptron (MLP) with 3 layers, which projects all agent-centric polylines to vectors of dimension 256.
Afterwards, we add learned absolute positional embeddings \cite{devlin2019bert}.
The resulting vector sets per agent-centric view have different lengths depending on the complexity of the corresponding local map area.
Therefore, we use a learned reduction mechanism \cite{wagner2023redmotion} to transform these agent-centric views into fixed-size agent-centric embeddings (see $E_i$ in Fig.\,1).
Our reduction decoder includes 4 blocks of alternating self-attention, cross-attention, LayerNorm, and FeedForward layers. 
The self-attention blocks mix features within $V_i$, and the cross-attention blocks extract features from $V_i$ into learned road environment descriptor (RED) tokens $R_j$.
Consequently, each $E_i$ embedding has the same fixed size of 128 tokens as the set of learned RED tokens.

\textbf{Scene-wide modules:} To form a scene-wide embedding, we concatenate the agent-centric embeddings with global reference tokens $G_i$ and rearrange them.
$G_i$ tokens are learned as linear projections of rotation and translation vectors for agent-centric views into a common scene-wide reference frame.
Our rearranging changes the batch dimension from the agent index to the scene index.
Therefore, our model has a dynamic batch size that changes between our agent-centric and scene-wide modules.
Afterwards, our latent context module learns global context via 6 blocks of self-attention, LayerNorm, and FeedForward layers.
Finally, we use an attention-based motion decoder with an MLP-based head to transform learned anchors $A_k$ into $k = 6$ motion modes.

\begin{table*}[b]
\setlength{\tabcolsep}{10pt}
\centering
\label{tab:1}
\resizebox{0.9\textwidth}{!}{
    \begin{tabular}{lllccccc}
    \toprule
    Split & Method (config) & Venue & mAP\,$\uparrow$ & minSADE\,$\downarrow$ & minSFDE\,$\downarrow$ & MR\,$\downarrow$ & OR\,$\downarrow$ \\ \midrule
    \multirow{5}{*}{Test} & Scene Transformer (joint) \cite{ngiam2022scene} & ICLR’22 & 0.1192 & \underline{0.9774} & \underline{2.1892} & \underline{0.4942} & \textbf{0.2067} \\
         & M2I \cite{sun2022m2i} & CVPR'22 & 0.1239 & 1.3506 & 2.8325 & 0.5538 & 0.2757 \\
         & GameFormer (joint) \cite{huang2023gameformer} & ICCV’23 & \underline{0.1376} & \textbf{0.9161} & \textbf{1.9373} & \textbf{0.4531} & \underline{0.2112} \\
         & SceneMotion (ours) & & \textbf{0.1789} & 1.0044 & 2.3141 & 0.5331 & 0.2163 \\
         \cmidrule{2-8}
         & MotionLM (ensemble) & ICCV'23 & 0.2178 & 0.8911 & 2.0067 & 0.4115 & -\\
         \midrule
    \multirow{5}{*}{Val}  & GameFormer (joint) \cite{huang2023gameformer} & ICCV'23 & 0.1339 & \textbf{0.9133} & \textbf{1.9251} & \underline{0.4564} & - \\
         & MotionLM (single replica) \cite{seff2023motionlm} & ICCV’23 & 0.1687 & 1.0345 & 2.3886 & 0.4943 & - \\
         & JFP (Wayformer) \cite{luo2023jfp} & CoRL'23 & \underline{0.1700} & \underline{0.9500} & \underline{2.1700} & \textbf{0.4500} & - \\
         & SceneMotion (ours) & & \textbf{0.1793} & 0.9613 & 2.2085 & 0.5137 & 0.2129 \\
         \cmidrule{2-8}
         & Wayformer (marginal) \cite{nayakanti2023wayformer, luo2023jfp} & ICRA'23 & 0.1600 & 0.9900 & 2.3000 & 0.4700 & - \\
         \bottomrule
    \end{tabular}
}
\caption{\textbf{Comparison of joint motion forecasting methods.} All methods are evaluated on the interactive splits of the Waymo Open Motion \mbox{Dataset \cite{ettinger2021large}}. Following \cite{zhang2023real}, we compare our method with real-time capable single replica versions of recent motion forecasting methods. We show the results of a non-realtime capable ensemble and a state-of-the-art marginal forecasting model for reference. Best scores are \textbf{bold}, second best are \underline{underlined}.}
\end{table*}

\textbf{Output representation:} We follow common practice (e.g., \cite{nayakanti2023wayformer, zhang2023real}) and represent future motion as mixture of Gaussians, with the following mean and covariance parameters for 2D positions: $\mu_x, \mu_y, \sigma_x, \sigma_y, \rho$. 
We forecast motion with a sampling frequency of 10\,Hz.
During training, we minimize the negative log-likelihood (NLL) loss for predicted positions and the cross-entropy loss for the corresponding confidence scores.
To learn joint modes of scene-wide forecasts, we use a winner-takes-all assignment to select the best of 6 scene-wide trajectory sets.
Specifically, we only backpropagate the loss for the set with the lowest average displacement error \cite{gupta2018social, ngiam2022scene}.

\section{Benchmarking motion forecasts}
In this section, we compare our method to recent motion forecasting methods using the large-scale Waymo Open Motion Dataset \cite{ettinger2021large}.
We focus our comparison on joint motion forecasting, but discuss marginal performance as well. 

\textbf{Dataset:} The Waymo Open Motion Dataset includes over 1.1 million data points extracted from 103,000 urban or suburban driving scenarios, spanning 20 seconds each.
The state of traffic agents includes attributes like class (i.e., vehicle, pedestrian, or cyclist), position, dimensions, velocity, acceleration, orientation, angular velocity, and the status of turn signals.
Each data point captures 1 second of past followed by 8 seconds of future data.
We resample this time interval with 10\,Hz.
The map data is represented as polylines and includes multiple lane types, crosswalks, speed bumps, stop signs, and traffic lights with their states. 

\textbf{Training details:} We configure our model as described in \cref{sec:method}, which corresponds to a transformer model with 19\,M trainable parameters.
We sample 28 scenes in a batch, with 8 focal (i.e., predicted) agents per scene.
Consequently, our agent-centric modules are trained with a batch size of 224 and our scene-wide modules with a batch size of 28.
We use Adam with weight decay \cite{loshchilov2018decoupled} as the optimizer and a step learning rate scheduler to halve the initial learning rate of $2^{-4}$ every 20 epochs.
We train for 50 epochs using data distributed parallel (DDP) training on 4 A6000 GPUs.

\textbf{Evaluation metrics:} We use the official challenge metrics, the mean average precision (mAP), the average displacement error (minADE), and the final displacement error (minFDE), the miss rate (MR), and the overlap rate (OR) to evaluate motion forecasts.
All metrics are computed using the minimum mode for $k = 6$ modes.
Accordingly, the metrics for the mode closest to the ground truth are measured.
Joint modes (i.e., best scene-wide mode) are evaluated on the interactive splits (denoted by an additional S in the metric names) and marginal modes (i.e., best mode for each agent individually) on the regular splits.
All metrics are averaged over the three prediction horizons of 3\,s, 5\,s, and 8\,s, and the 3 agent classes (vehicles, pedestrians, and cyclists).
Refer to the appendix for the complete results of our model.  

\textbf{Joint motion forecasting performance:} Table I shows the results of our method on the official Waymo Interaction Prediction challenge. 
Therefore, all metrics are scene-wide and computed for joint modes. 
Following \cite{zhang2023real}, we compare against real-time capable single replica versions of joint motion forecasting methods and marginal methods with auxiliary joint motion forecasting objectives.
Additionally, we show the results of a non-realtime capable ensemble of 8 MotionLM models and a state-of-the-art marginal motion forecasting model (Wayformer) for reference.
In both splits (test and validation), our method outperforms all comparable methods on the main challenge metric, the mAP. 
In particular, our model outperforms a single replica of MotionLM; only a non-real-time capable ensemble of 8 replicas achieves higher mAP values on the test set.
For most of the remaining metrics, our model ranks third.
\begin{figure*}[h!]
    \centering
    \vspace{2mm}
    \includegraphics[width=\textwidth]{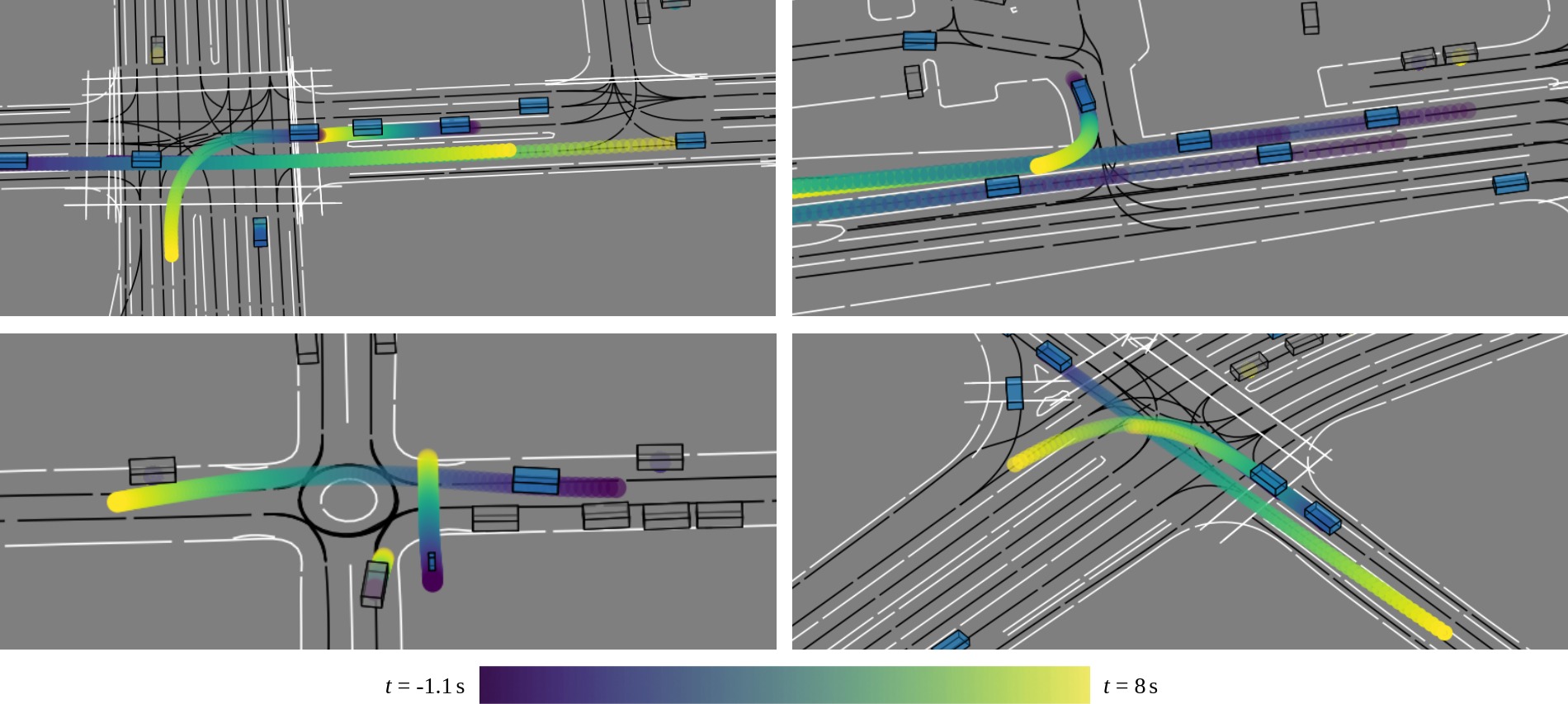}
    \caption{\textbf{Scene-wide motion forecasts.} Our model forecasts scene-wide motion modes by modeling joint distributions of trajectories for 8 focal agents. Dynamic agents are shown in blue, static agents in grey (determined at $t=0\,\text{s}$). Lanes are black lines and road markings are white lines.}
    \label{fig:enter-label}
\end{figure*}
Furthermore, all joint models except GameFormer outperform the marginal Wayformer model, underscoring the importance of joint motion forecasting for interaction modeling. 
Fig.\,2 shows qualitative results from the interactive split of the Waymo Open Motion dataset.

\textbf{Marginal motion forecasting performance:} Our model is flexible with respect to forecasting types and can be configured to forecast marginal distributions of trajectories (i.e., for each agent independently).
Specifically, we fine-tune our previously trained model for marginal forecasting by changing the winner-takes-all assignment to choose the best mode per agent instead of the best scene-wide mode.
We fine-tune for another 40 epochs, for a total of 90 epochs, and keep the training hyperparameters the same as in the initial training. 
Table II shows that our model achieves competitive results on marginal motion forecasting metrics as well.

\begin{table*}[ht!]
\setlength{\tabcolsep}{10pt}
\centering
\label{tab:2}
\resizebox{0.9\textwidth}{!}{
    \begin{tabular}{llcccccc}
    \toprule
     Method (config) & Venue & mAP\,$\uparrow$ & minADE\,$\downarrow$ & minFDE\,$\downarrow$ & \#Epochs & $T_{\text{epoch}}$ & $T_{\text{train}}$\\ \midrule
    Scene Transformer (marginal) \cite{ngiam2022scene} & ICLR’22 & - & 0.6130 & 1.2200 & - & - & - \\
          MTR \cite{shi2022motion} & NeurIPS'22 & \underline{0.4164} & 0.6046 & 1.2251 &  - & - & - \\
          HPTR \cite{zhang2023real} & NeurIPS'23 & 0.4150 & \textbf{0.5378} & \textbf{1.0923} & 120 & 52\,min & 104\,h \\
          SceneMotion (ours, marginal) & & \textbf{0.4251} & \underline{0.5464} & \underline{1.1202} & 90 & 22\,min & 33\,h \\
         \bottomrule
    \end{tabular}
}
\caption{\textbf{Marginal motion forecasting performance and training time.} All methods are evaluated on the regular validation split of the Waymo Open Motion Dataset \cite{ettinger2021large}. $T_{\text{epoch}}$ and $T_{\text{train}}$ are measured for training runs with 4 A6000 GPUs. Best scores are \textbf{bold}, second best are \underline{underlined}.}
\end{table*}

\textbf{Training time:}
We compare the required training epochs with those of the recent pairwise relative model HPTR. 
Our two-stage training (first joint, then marginal) is 90 epochs long, compared to the 120 epochs recommended for HPTR. 
Since our model allows for larger batch sizes on the same hardware, the time required to train one epoch is more than 50\% shorter.
As a result, the total training time $T_{\text{train}}$ of our model is only about one third of the time required to train HPTR using A6000 GPUs (see Table II).

\section{Analyzing clusters of future waypoints} 
In this section, we propose a method to quantify future interaction based on our multimodal motion forecasts.
Our method clusters future waypoints to determine which agents are likely to interact in the future.
We compute clusters for all agents over all modes in each timestep (i.e., we initially discard motion modes by merging trajectories from all modes).
Afterwards, we evaluate if our forecasting model resolves the waypoint clusters by modeling the interaction between agents.
In particular, we verify that the clusters do not exist within joint modes.
Vice versa, if the cluster exists within a joint motion mode, a future conflict is more likely.
Thus, in subsequent planning, trajectories with waypoint clusters can be associated with higher risk \cite{hubmannAutomatedDrivingUncertain2018, tas2022motion, huang2024learning}.

Specifically, we use the DBSCAN algorithm \cite{ester1996density} for clustering, with a max. distance of 2.5\,m and a min. number of 2 waypoints per cluster.
Unlike clustering algorithms that require specifying the number of clusters beforehand (e.g., $k$-means), DBSCAN clusters waypoints more reliably by choosing a suitable number of clusters depending on the given data.

For illustration, we show an example of two interacting agents in Fig.\;3, yet our method scales to an arbitrary number of agents.
In Fig.\;3\,(a), waypoints of the lower vehicle that form clusters with waypoints of another vehicle are highlighted in orange.
The vehicle potentially involved in the interaction is connected to these waypoints by dashed orange lines.
Fig.\;3\,(b) shows all trajectories for the corresponding vehicles leading to the waypoint clusters.
Since our model is trained to forecast consistent modes with respect to the future motion of agents, waypoint clusters are less likely within joint modes.
For example, Fig.\;3\,(c) shows the trajectories of the involved vehicles from the top-1 (i.e., most likely) joint motion mode, which does not include a waypoint cluster.
Thus, our forecasting model likely resolves the cluster by modeling the interaction between the two vehicles. 

\begin{figure}[h!]
    \centering
    \vspace{2mm}
    \includegraphics[width=0.49\textwidth]{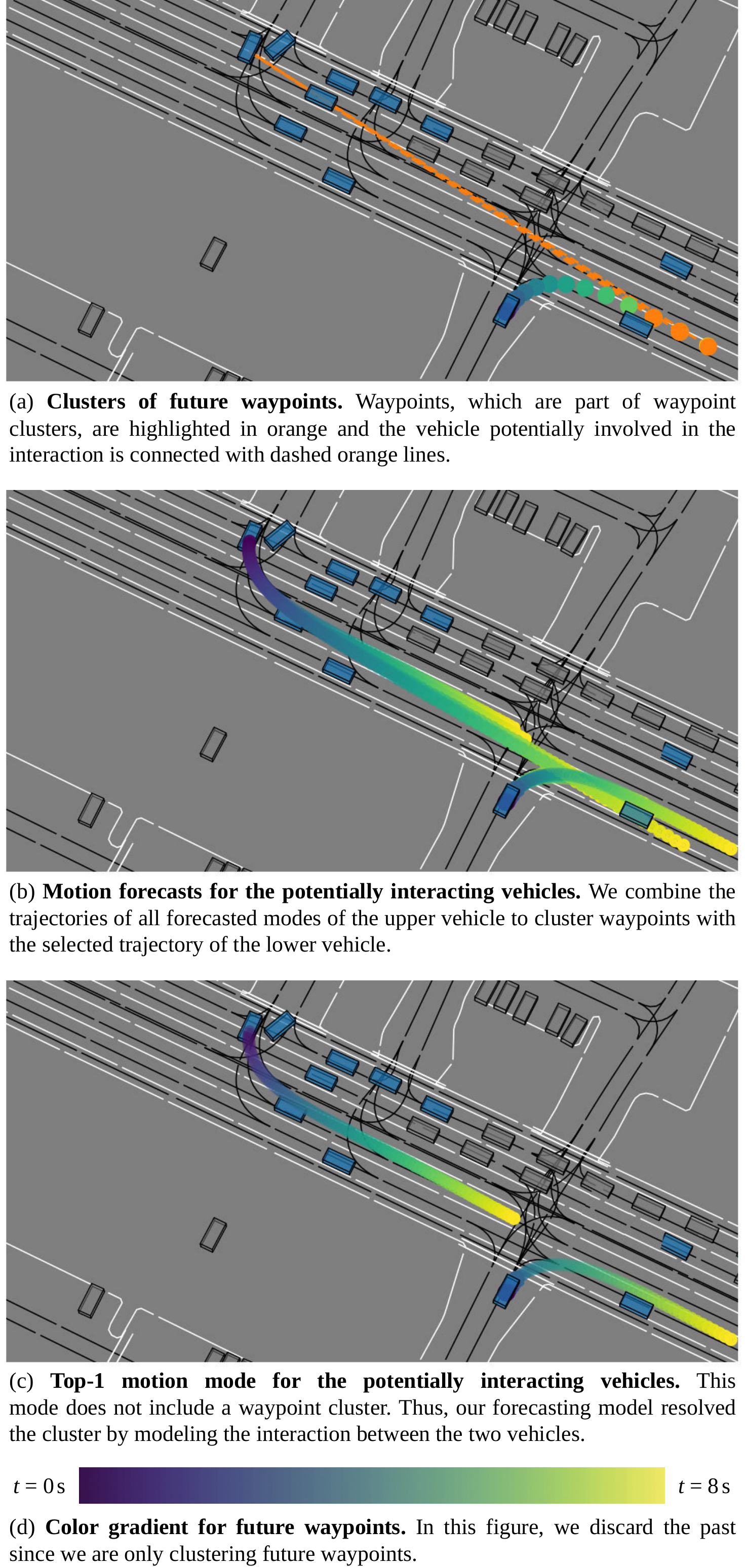}
    \caption{\textbf{Analyzing scene-wide motion forecasts in terms of waypoint clusters.} We show an example of a potential interaction of widely separated vehicles to demonstrate the benefits of a long prediction horizon of 8\,s.}
    \label{fig:enter-label}
\end{figure}

We evaluate our method on the interactive validation split of the Waymo Open Motion Dataset.
The results are shown in Table\,III. 
When clustering the merged set of waypoints from all motion modes, $29.42\%$ of the agents have clustered waypoints with other agents.
This shows a large number of possible future interactions.

Notably, the percentage of agents with clustered waypoints decreases to $13.52\%$ when all modes are considered independently and to $6.55\%$ when only the top-1  mode is considered.

Furthermore, we evaluate the learned assignment of trajectories to scene-wide modes by comparing it to a random assignment scheme.
Specifically, we compute the average percentage of agents with clustered waypoints within the forecasted modes versus with random mode assignment per trajectory.
We implement the random assignment scheme by sampling mode indices for each trajectory from a discrete uniform distribution on the integer interval $[0\,..\,5]$.
\mbox{Within the resulting} modes of randomly shuffled trajectories, we compute the percentage of agents with clustered waypoints.
We repeat this process 6 times per scene and take the average (see ``Avg random mode per trajectory'' in Table III).
The average percentage of agents with clustered waypoints is 23\% lower within the forecasted modes (i.e., using the learned assignment) than with random assignment (6.65\% versus 8.69\%). 

Overall, the results indicate that our motion forecasting model is capable of interaction modeling and resolving potential future conflicts.

\begin{table}[h!]
\centering
\label{tab:conflicts}
\begin{tabular}{l r}
\toprule
\text{Evaluation type} & \text{Clustered agents} $\downarrow$ \\ \midrule
\text{All modes merged} & $29.42$\% \\
\text{Top-1 mode} & $6.55$\% \\
\text{Top-3 modes} & $10.12$\% \\
\text{Top-6 modes} & $13.52$\% \\ 
\midrule
Avg random mode per trajectory & 8.69\% \\
Avg within modes & 6.65\% \\
\bottomrule
\end{tabular}
\caption{\textbf{Percentage of agents with clustered waypoints.} ``All modes merged'' refers to clustering waypoints from all modes, while ``Top-$k$ modes'' refers to clustering within $k$ modes independently.
``Avg random mode per trajectory'' refers to the average of 6 iterations of random sampling of mode indices per trajectory. ``Avg within modes'' refers to the average of the 6 forecasted modes using the learned mode indices. 
}
\end{table}

\section{Conclusion and future work}
\label{sec:conclusion}

In this work, we have introduced a method for transforming local agent-centric embeddings into scene-wide motion forecasts.
Inspired by methods that perform self-supervised pre-training, we use local agent-centric views for more data-efficient learning compared to existing scene-centric and pairwise relative representations.
Our novel latent context module subsequently learns a scene-wide latent space from multiple agent-centric embeddings, enabling joint motion forecasting and interaction modeling.
Evaluations on the Waymo Open Motion Dataset and the Waymo Open \mbox{Interaction} Prediction Challenge confirm the effectiveness of our method.

Furthermore, we have proposed to cluster future waypoints in time and space to quantify the interaction between agents.We merge all modes and analyze each mode independently to determine whether a cluster is likely resolved by interaction or results in conflict.

Future work includes exploring how motion planners can benefit from additional information about interactions and the use of mixture models for specific agents and scenarios.

\section*{Acknowledgments}
We acknowledge the financial support for this work by the Federal Ministry of Education and Research of Germany (BMBF) within the projects HAIBrid (FKZ 01IS21096A) and AUTOtech.agil (FKZ 01IS22088T).
Furthermore, this work was supported by the Helmholtz Association's Initiative and Networking Fund on the HAICORE@FZJ partition.

\clearpage
\printbibliography

\newpage

\begin{samepage}
\begin{textblock*}{\textwidth}(95mm,20mm)
    \textsc{Appendix}
\end{textblock*}

\vspace{20mm}
\begin{table*}
\vspace{5mm}
\centering
\captionsetup{justification=centering} 
\resizebox{0.75\textwidth}{!}{
\begin{tabular}{@{}lccccccc@{}}
\toprule
Object Type & Measurement time (s) & soft mAP\,$\uparrow$ & mAP\,$\uparrow$ & minADE\,$\downarrow$ & minFDE\,$\downarrow$ & MR\,$\downarrow$ & OR\,$\downarrow$ \\ \midrule
Vehicle     & 3                    & 0.3649   & 0.3466 & 0.4016 & 0.8177 & 0.2962 & 0.0421 \\
Vehicle     & 5                    & 0.2405   & 0.2295 & 0.8685 & 1.9423 & 0.4363 & 0.1245 \\
Vehicle     & 8                    & 0.1371   & 0.1319 & 1.8969 & 4.6197 & 0.6019 & 0.3184 \\
Vehicle     & Avg                  & 0.2475   & 0.2360 & 1.0557 & 2.4599 & 0.4448 & 0.1616 \\
Pedestrian  & 3                    & 0.3218   & 0.3104 & 0.3128 & 0.6283 & 0.3646 & 0.2195 \\
Pedestrian  & 5                    & 0.2118   & 0.2040 & 0.6669 & 1.4729 & 0.5170 & 0.2959 \\
Pedestrian  & 8                    & 0.1352   & 0.1346 & 1.4081 & 3.3252 & 0.6932 & 0.4419 \\
Pedestrian  & Avg                  & 0.2229   & 0.2164 & 0.7959 & 1.8088 & 0.5250 & 0.3191 \\
Cyclist     & 3                    & 0.1567   & 0.1512 & 0.4755 & 0.9547 & 0.5015 & 0.0673 \\
Cyclist     & 5                    & 0.0759   & 0.0736 & 0.9836 & 2.1837 & 0.6148 & 0.1498 \\
Cyclist     & 8                    & 0.0284   & 0.0280 & 2.0260 & 4.8828 & 0.7725 & 0.2876 \\
Cyclist     & Avg                  & 0.0870   & 0.0843 & 1.1617 & 2.6737 & 0.6296 & 0.1682 \\
All         & 3                    & 0.2812   & 0.2694 & 0.3967 & 0.8002 & 0.3874 & 0.1096 \\
All         & 5                    & 0.1761   & 0.1691 & 0.8397 & 1.8663 & 0.5227 & 0.1901 \\
All         & 8                    & 0.1002   & 0.0982 & 1.7770 & 4.2759 & 0.6892 & 0.3493 \\
All         & Avg                  & 0.1858   & 0.1789 & 1.0044 & 2.3141 & 0.5331 & 0.2163 \\ \bottomrule
\end{tabular}
}
\centering
\caption{Test metrics by object type and measurement time for joint motion forecasting [\href{https://waymo.com/open/challenges/interaction-prediction/results/824a6c20-3f34/1713653531162000}{challenge website}].
According to the definition of \cite{zhang2023real}, our models are end-to-end since we do not use ensembling techniques with multiple models or multiple decoders, and our models forecast 6 trajectories, which does not require trajectory aggregation as post-processing.}
\end{table*}

\begin{table*}[ht!]
\centering
\resizebox{0.75\textwidth}{!}{
\begin{tabular}{@{}lccccccc@{}}
\toprule
Object Type & Measurement time (s) & soft mAP\,$\uparrow$ & mAP\,$\uparrow$ & minADE\,$\downarrow$ & minFDE\,$\downarrow$ & MR\,$\downarrow$ & OR\,$\downarrow$ \\  \midrule
Vehicle     & 3                    & 0.3835   & 0.3678 & 0.3953 & 0.8006 & 0.2864 & 0.0396 \\
Vehicle     & 5                    & 0.2622   & 0.2543 & 0.8491 & 1.8859 & 0.4225 & 0.1205 \\
Vehicle     & 8                    & 0.1788   & 0.1757 & 1.8448 & 4.4780 & 0.5906 & 0.3132 \\
Vehicle     & Avg                  & 0.2748   & 0.2659 & 1.0297 & 2.3882 & 0.4332 & 0.1578 \\
Pedestrian  & 3                    & 0.3070   & 0.2944 & 0.2990 & 0.5890 & 0.3542 & 0.2185 \\
Pedestrian  & 5                    & 0.1946   & 0.1896 & 0.6175 & 1.3270 & 0.4681 & 0.2928 \\
Pedestrian  & 8                    & 0.0725   & 0.0691 & 1.2874 & 2.9752 & 0.6350 & 0.4306 \\
Pedestrian  & Avg                  & 0.1914   & 0.1844 & 0.7346 & 1.6304 & 0.4858 & 0.3139 \\
Cyclist     & 3                    & 0.1624   & 0.1579 & 0.4582 & 0.9194 & 0.4974 & 0.0688 \\
Cyclist     & 5                    & 0.0774   & 0.0760 & 0.9431 & 2.0950 & 0.6151 & 0.1436 \\
Cyclist     & 8                    & 0.0288   & 0.0285 & 1.9576 & 4.8060 & 0.7536 & 0.2884 \\
Cyclist     & Avg                  & 0.0895   & 0.0875 & 1.1196 & 2.6068 & 0.6221 & 0.1669 \\
All         & 3                    & 0.2843   & 0.2734 & 0.3842 & 0.7697 & 0.3793 & 0.1090 \\
All         & 5                    & 0.1781   & 0.1733 & 0.8032 & 1.7693 & 0.5019 & 0.1856 \\
All         & 8                    & 0.0933   & 0.0911 & 1.6966 & 4.0864 & 0.6598 & 0.3441 \\
All         & Avg                  & 0.1852   & 0.1793 & 0.9613 & 2.2085 & 0.5137 & 0.2129 \\ \bottomrule
\end{tabular}
}
\captionsetup{justification=centering}
\centering
\caption{Validation metrics by object type and measurement time for joint motion forecasting [\href{https://waymo.com/open/challenges/interaction-prediction/results/824a6c20-3f34/1713650971850000}{challenge website}].}
\end{table*}

\begin{table*}
\centering
\captionsetup{justification=centering} 
\resizebox{0.75\textwidth}{!}{
\begin{tabular}{@{}lccccccc@{}}
\toprule
Object Type & Measurement time (s) & soft mAP\,$\uparrow$ & mAP\,$\uparrow$ & minADE\,$\downarrow$ & minFDE\,$\downarrow$ & MR\,$\downarrow$ & OR\,$\downarrow$ \\ \midrule
Vehicle & 3 & 0.5721 & 0.5540 & 0.2819 & 0.4909 & 0.0885 & 0.0184 \\
Vehicle & 5 & 0.4669 & 0.4598 & 0.5721 & 1.1001 & 0.1271 & 0.0391 \\
Vehicle & 8 & 0.3483 & 0.3450 & 1.1030 & 2.4312 & 0.2019 & 0.0940 \\
Vehicle & Avg & 0.4625 & 0.4529 & 0.6523 & 1.3407 & 0.1392 & 0.0505 \\
Pedestrian & 3 & 0.5566 & 0.5360 & 0.1494 & 0.2709 & 0.0505 & 0.2383 \\
Pedestrian & 5 & 0.4653 & 0.4562 & 0.2821 & 0.5517 & 0.0687 & 0.2643 \\
Pedestrian & 8 & 0.4180 & 0.4109 & 0.4896 & 1.0412 & 0.0861 & 0.2962 \\
Pedestrian & Avg & 0.4800 & 0.4677 & 0.3071 & 0.6213 & 0.0684 & 0.2662 \\
Cyclist & 3 & 0.4464 & 0.4341 & 0.3356 & 0.6049 & 0.1900 & 0.0451 \\
Cyclist & 5 & 0.3608 & 0.3570 & 0.6174 & 1.1907 & 0.1969 & 0.0842 \\
Cyclist & 8 & 0.2746 & 0.2727 & 1.0859 & 2.3999 & 0.2458 & 0.1389 \\
Cyclist & Avg & 0.3606 & 0.3546 & 0.6797 & 1.3985 & 0.2109 & 0.0894 \\
All & 3 & 0.5250 & 0.5080 & 0.2556 & 0.4556 & 0.1097 & 0.1006 \\
All & 5 & 0.4310 & 0.4243 & 0.4905 & 0.9475 & 0.1309 & 0.1292 \\
All & 8 & 0.3469 & 0.3429 & 0.8929 & 1.9574 & 0.1779 & 0.1764 \\
All & Avg & 0.4343 & 0.4251 & 0.5464 & 1.1202 & 0.1395 & 0.1354 \\
\bottomrule
\end{tabular}
}
\caption{Validation metrics by object type and measurement time for marginal motion forecasting [\href{https://waymo.com/open/challenges/motion-prediction/results/824a6c20-3f34/1714402734175000}{challenge website}].}
\end{table*}

\begin{table*}[ht]
  \centering
  \label{tab:performance}
  \begin{tabular}{lc}
    \toprule
    Focal agents & Inference latency \\
    \midrule
     2 & \,\,\,54.3\,ms \\
     8 (default) & \,\,\,59.7\,ms \\
     64 &  191.1\,ms\\
    \bottomrule
  \end{tabular}
\caption{\textbf{Inference latency of our SceneMotion model.} We measure the inference latency on one A6000 GPU using the PyTorch Lightning profiler and plain eager execution. We report the average of 1000 iterations per configuration for the complete \texttt{predict\_step}, including pre- and post-processing.}
\end{table*}

\end{samepage}
\FloatBarrier

\addtolength{\textheight}{-12cm}
\end{document}